\documentclass[preprint,12pt]{elsarticle}

\usepackage{amssymb}
\usepackage{amsmath}
\usepackage{subcaption}

\journal{Nuclear Physics B}
\usepackage{lineno}
\begin{document}

\begin{frontmatter}



\title{Wav2Arrest 2.0: Long-Horizon Cardiac Arrest Prediction with Time-to-Event Modeling, Identity-Invariance, and Pseudo-Lab Alignment}


\author[1]{Saurabh Kataria} 
\author[1]{Davood Fattahi}
\author[1]{Minxiao Wang}
\author[1]{Ran Xiao}
\author[2]{Matthew Clark}
\author[2]{Timothy Ruchti}
\author[3,4]{Mark Mai}
\author[1]{Xiao Hu}





\address[1]{Nell Hodgson Woodruff School of Nursing, Emory University, 1520 Clifton Road NE, Atlanta, GA 30322, USA}
\address[2]{Nihon Kohden Digital Health Solutions, 15353 Barranca Parkway, Irvine, CA 92618, USA}
\address[3]{Children's Healthcare of Atlanta, 2220 N Druid Hills Rd NE, Atlanta, GA 30329, USA}
\address[4]{Department of Pediatrics, Emory School of Medicine, 100 Woodruff Circle, Atlanta, GA 30322, USA}

\begin{abstract}
High-frequency physiological waveform modality offers deep, real-time insights into patient status.
Recently, physiological foundation models based on Photoplethysmography (PPG), such as PPG-GPT, have been shown to predict critical events, including Cardiac Arrest (CA).
However, their powerful representation still needs to be leveraged suitably, especially when the downstream data/label is scarce.
We offer three orthogonal improvements to improve PPG-only CA systems by using minimal auxiliary information.
First, we propose to use time-to-event modeling, either through simple regression to the event onset time or by pursuing fine-grained discrete survival modeling.
Second, we encourage the model to learn CA-focused features by making them patient-identity invariant.
This is achieved by first training the largest-scale de-identified biometric identification model, referred to as the p-vector, and subsequently using it adversarially to deconfound cues, such as person identity, that may cause overfitting through memorization.
Third, we propose regression on the pseudo-lab values generated by pre-trained auxiliary estimator networks.
This is crucial since true blood lab measurements, such as lactate, sodium, troponin, and potassium, are collected sparingly.
Via zero-shot prediction, the auxiliary networks can enrich cardiac arrest waveform labels and generate pseudo-continuous estimates as targets.
Our proposals can independently improve the 24-hour time-averaged AUC from the 0.74 to the 0.78-0.80 range.
We primarily improve over longer time horizons with minimal degradation near the event, thus pushing the Early Warning System research.
Finally, we pursue multi-task formulation and diagnose it with a high gradient conflict rate among competing losses, which we alleviate via the PCGrad optimization technique.
\end{abstract}



\begin{keyword}
Cardiac Arrest, Time-to-Event, Patient Biometric, Pseudo Laboratory Values, Physiological Foundation Model



\end{keyword}

\end{frontmatter}


\section{Introduction}
Predicting critical events and associated patient deterioration is an important topic in Artificial Intelligence (AI)-based critical care~\cite{pinsky2024use}.
To facilitate timely clinical intervention, it is even crucial to \emph{predict several hours ahead of the event onset} -- which is referred to as long-horizon prediction.
Examples of such events include cardiac arrest~\cite{kataria2025continuous,lee2024prediction,lee2023real}, myocardial infarction~\cite{gustafsson2022development}, sepsis~\cite{lombardi2022classifying}, and respiratory decline~\cite{spijkerboer2024machine}, among others.
The prediction systems may be deployed by generating \emph{alarms} and \emph{warnings} for clinicians.
Specifically, Early Warning Systems (EWS) are developed, and they are typically multimodal -- often combining variables from the Electronic Health Record (EHR) with patient measurements from devices.
Clinical risk scores, such as the Modified Early Warning Score (MEWS)~\cite{tan2022modified} and Electronic Cardiac Arrest Risk Triage (eCART)~\cite{bartkowiak2019validating}, utilize additional laboratory reports and EHR information.
Custom warning systems for specific abnormalities may also be developed.
For instance, in \cite{sundrani2023predicting}, the authors combined multi-modal triage data and brief physiological signals to predict vital sign abnormalities.
We argue that developing unimodal risk assessment models, based on device measurements, is equally essential for several reasons, including addressing missing modality, modality fusion, and edge device deployment, among others.  
Moreover, device-derived data are continuous and not subject to the potential bias inherent in clinician-ordered measurements.

EWS systems are now utilizing robust features from Foundation Models (FMs), which are trained on large-scale data.
Physiological waveform FMs like SiamQuality~\cite{ding2024siamquality}, PPG-GPT~\cite{chen2024adapting,chen2025gpt}, PaPaGei~\cite{pillai2024papagei} and Pulse-PPG~\cite{saha2025pulse} are PPG modality-specific models.
Multi-modal sensor models, such as LSM~\cite{narayanswamy2024scaling} and NormWear~\cite{luo2024toward}, utilize multiple sensor signals and proprietary feature extractors to train large-scale foundation models.
There are also specialized Intensive Care Unit (ICU) FMs, such as ICareFM~\cite{burger2025foundation}, which harness vast amounts of irregular data to train with a Time-to-Event (TTE) objective.

In \cite{kataria2025continuous,kataria2025cardiac}, the authors proposed the Feature Extractor-Aggregator Network (FEAN), which utilizes PPG FMs for cardiac arrest forecast up to 24 hours in advance.
Our new proposal, Wav2Arrest 2.0 model (W2A2), is an improvement over the FEAN model by introducing three orthogonal ideas.
First, we propose incorporating TTE, either through regression on event time or by employing a survival model.
TTE can leverage the cardiac arrest event time label, which is typically available.
In \cite{lin2025ecg}, the authors used a 12-lead ECG to predict 1-year mortality.
However, we focus on shorter horizons as done in continuous monitoring~\cite{kataria2025continuous} in the ICU.
Second, we introduce a mechanism to prevent the cardiac arrest model from memorizing information, such as patient identity, demographics, and hospital- and treatment-specific details.
This focuses the model to learn generalizable markers of cardiac arrest.
Adversarial learning~\cite{ganin2016domain} enables us to deconfound patient identity by utilizing a pre-trained, large-scale patient identification model, which we refer to as a p-vector.
Third, we align/ground learned features with evolving physiology via regression to pseudo-lab values -- leading to a Multi-Task Learning (MTL) setup.
We simulate such longitudinal labels, as the true lab values are expensive to obtain and sparsely collected as per standard clinical practice.
We simulate using auxiliary networks that are pre-trained on blood biomarker estimation, which is supported by prior work~\cite{wang2025estimatingclinicallabtest}.
In a relevant Sepsis prediction work~\cite{roussel2019recurrent}, the authors pursued MTL of Sepsis with future vital prediction using input consisting of current vital signs, laboratory results, and demographics information — in contrast to our work's focus on PPG input only.

In this study, we
1) Propose three orthogonal ideas to improve waveform-only cardiac arrest prediction and compare the baseline with different foundation models.
2) Compare simple Time-to-Event regression with survival modeling and improve predictions over a long horizon.
3) Emphasize the role of patient identity invariance and implement it with the largest de-identified patient biometric model.
4) Align learned features with the estimated physiology trajectory by leveraging pre-trained blood biomarker estimator networks.
5) Present a comprehensive gradient conflict analysis in multi-task learning and its partial mitigation via PCGrad.

\section{Baseline FEAN model}
\begin{figure*}[!t]
  \centering
  \includegraphics[width=0.99\textwidth]{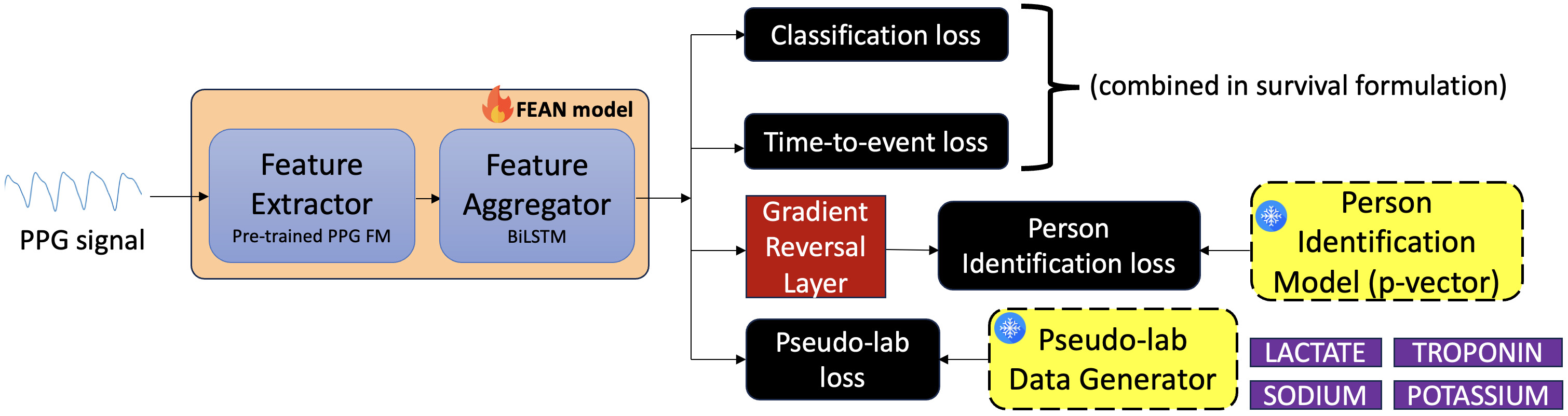}
  \caption{Overview of Wav2Arrest 2.0 (W2A2) training framework.
  A PPG waveform is processed by FEAN (a PPG foundation model feature extractor followed by a BiLSTM aggregator) to produce a representation $\mathbf{z}$. 
  The proposed auxiliary terms include 1) time-to-event modeling (standalone as regression or with main classification loss in hazard formulation), 2) patient-identity-invariant representation learning via gradient reversal and pretrained identity (p-vector) model, 3) pseudo-lab prediction loss via pre-trained pseudo-lab generator model.}
  \label{fig:wav2arrest2}
\end{figure*}

In \cite{kataria2025continuous}, the authors proposed an automated cardiac arrest prediction model called FEAN (Feature Extractor-Aggregator Network) using just Photoplethysmography signals and corresponding cardiac arrest binary labels.
For illustration, consider the first half of Fig.~\ref{fig:wav2arrest2} with just the classification loss.
The first block is the \emph{feature extractor}, for which we use time-series (both generalist and clinical) Foundation Models (FM) that the original study did not explore.
We choose a small variant of PPG-trained PaPaGei and generalist MOMENT for comparison.
For feature extraction, we independently extract a representative embedding from each 30-second segment.
The second block is the \emph{feature aggregator} for which the prior work identified BiLSTM (Bidirectional Long-Short Term Memory)~\cite{hochreiter1997long} network to be a robust choice.

We follow the same experimental setup as FEAN.
Due to the computational complexity of our proposal, we chose the small variants of all FMs.
We applied the approach to the UCSF ICU dataset~\cite{drew2014insights,ding2024siamquality}, which is filtered down to 200 case (positive) patients and 1000 control (negative) patients.
Among all signals and the corresponding Electronic Health Record (EHR) information, only the PPG signals were used.
For case patients, data from 25 hours (t=T-25) to 1 hour (t=T-1) before the event onset time (t=T) were extracted.
For controls, the last 24 hours of their stay were extracted.
This may correspond to the patient's recovery phase; however, we will conduct a thorough exploration of control selection in future work.
Additionally, we employ only the 1H model from FEAN, i.e., our system consumes the latest one hour of patient data history only to make predictions.
Optimization is performed using MARS~\cite{yuan2024mars} with a batch size of 256 and 50 epochs.
Due to the small size of the downstream tuning data and noisy labels, the results are sensitive to the choice of hyperparameters.
We fix the (learning rate, weight decay) for the feature extractor and the aggregator to (0.0002,0.001) and (0.00001,0), respectively.
Additionally, the data was stratified into training and testing sets with an 80:20 ratio to conduct the experiments.
FEAN trains with a simple Binary Cross-Entropy (BCE) loss for the classification problem.
Our proposal, Wav2Arrest 2.0 (W2A2), improves upon FEAN in three aspects of objective formulation as explained below.

\section{Wav2Arrest 2.0}
We propose independent additional loss terms to the regular BCE loss to learn more generalizable long-term markers of cardiac arrest.
Figure~\ref{fig:wav2arrest2} illustrates the complete model.
We now explain the three sub-parts of the proposal: 1) time-to-event modeling, 2) identity invariance, and 3) pseudo-lab targets.


\subsection{Time-to-event Modeling}
The exact timestamp for the cardiac arrest onset is utilized here.
In the simpler formulation, the estimation of \emph{time to cardiac arrest} is added as an additional regression loss term, but only for the case/positive patients.
The term is not used for control patients.
The mathematical formulation is detailed below.
For BCE, the numerically stable form from PyTorch is used ~\cite{paszke2019pytorch}.

\noindent\textbf{Setup 1.}
For a minibatch of size $B$, let $y_i\in\{0,1\}$, class logit $s_i\in\mathbb{R}$,
predicted time-to-event $\hat t_i\in\mathbb{R}$, and (for positives) true TTE $t_i\in\mathbb{R}_{\ge 0}$.
Define the positive mask $m_i=\mathbb{I}[y_i=1]$, the count $N_{+}=\sum_{i=1}^{B} m_i$, and a small $\varepsilon>0$.

\noindent\textbf{Classification (BCE with logits).}
\begin{equation}
\mathcal{L}_{\mathrm{cls}}
=\frac{1}{B}\sum_{i=1}^{B}\!\left(\log\!\big(1+\text{exp}(s_i)\big)-y_i s_i\right).
\end{equation}

\noindent\textbf{TTE regression (Positive-only masked MSE).}
\begin{equation}
\mathcal{L}_{\mathrm{reg}}
=\frac{1}{N_{+}}\sum_{i=1}^{B}
m_i\,\big ||\hat t_i - t_i\big||_1 .
\end{equation}

\noindent\textbf{Total loss.}
\begin{equation}
\mathcal{L}
=\mathcal{L}_{\mathrm{cls}}
+\lambda_{\mathrm{reg}}\,\mathcal{L}_{\mathrm{reg}} .
\end{equation}

This formulation is simple yet powerful, with the limitation of not being able to utilize the survival information of negative patients.
To improve this regression TTE, various error metrics were explored beyond L1 loss; however, Mean-Squared Error (MSE) and Huber loss did not yield benefits.

A more sophisticated formulation is now considered, which extends beyond the simpler TTE regression and implements a discrete survival model by dividing the prediction horizon/analysis window into slots and assigning a \emph{hazard} value for each positive and negative patient.

\noindent\textbf{Setup 2.}
Let a fixed grid of $L$ future time bins of width $\Delta t$ (in our case, $1$ hour) be indexed by $l\in\{1,\ldots,L\}$.
Given a PPG segment, the BLSTM produces a context vector $\mathbf{z}\in\mathbb{R}^D$ and a CA logit $s\in\mathbb{R}$.
We define discrete survival targets and an at-risk mask~\cite{chen2024introduction}:
\begin{equation}
y_l \in \{0,1\}, \quad
m_l \in \{0,1\},
\end{equation}
where $y_l=1$ indicates the event occurs in bin $l$, and $m_l=1$ indicates bin $l$ is observed.

\textbf{Hazard formulation.}
We place an additive, time-varying head on top of $\mathbf{z}$:
\begin{equation}
\eta_l \;=\; \mathbf{z}^\top \mathbf{W}_{:,l} + b_l,
\qquad
h_l \;=\; \sigma(\eta_l),
\end{equation}
with sigmoid $\sigma(\cdot)$ and learnable $\mathbf{W}\in\mathbb{R}^{D\times L}$, $\mathbf{b}\in\mathbb{R}^{L}$.


\textbf{Discrete survival loss (masked BCE).}
Right-censored bins are excluded by $m_l$, at-risk bins contribute a per-bin BCE:
\begin{equation}
\begin{aligned}
\mathcal{L}_{\mathrm{surv}}
&= - \frac{1}{\sum_{l=1}^{L} m_l}
   \sum_{l=1}^{L} m_l \Big[
   y_l \log \sigma(\eta_l) \\
&\qquad\qquad\quad
   + (1-y_l)\log\!\big(1-\sigma(\eta_l)\big)
   \Big].
\end{aligned}
\end{equation}

\textbf{Regularization of time-varying effects.}
We encourage smooth variation across time and parsimony across features:
\begin{align}
\mathcal{L}_{\mathrm{smooth}}
&=\sum_{l=1}^{L-1} \left\| \mathbf{W}_{:,l+1} - \mathbf{W}_{:,l} \right\|_2^2,\\
\mathcal{L}_{\mathrm{group}}
&=\sum_{k=1}^{D} \sqrt{ \sum_{l=1}^{L} W_{k,l}^2 + \epsilon }.
\end{align}

\textbf{Total loss.}
The network is trained end-to-end with
\begin{equation}
\begin{aligned}
\mathcal{L}_{\mathrm{total}}
&= \mathcal{L}_{\mathrm{cls}}
 + \lambda_{\mathrm{surv}}\,\mathcal{L}_{\mathrm{surv}} \\
&\quad + \lambda_{\mathrm{smooth}}\,\mathcal{L}_{\mathrm{smooth}}
 + \lambda_{\mathrm{group}}\,\mathcal{L}_{\mathrm{group}}.
\end{aligned}
\end{equation}

The hyperparameters $\lambda_{\mathrm{surv}}$, $\mathcal{L}_{\mathrm{smooth}}$, and $\mathcal{L}_{\mathrm{group}}$ are set to 1, 0.001, and 0.0001, respectively.



\subsection{Identity Invariance}
Here, we aim to improve the model's generalization by (adversarially) suppressing patient-specific information, thereby encouraging the learning of features related solely to the cardiac arrest task.
Inspired by the speaker recognition problem~\cite{mak2020machine}, speaker recognition model x-vector~\cite{snyder2018x} in audio research, we first train a large-scale patient identification model using 0.5 million hours of PPG derived from de-identified 7068 patients in MIMIC-III data~\cite{johnson2016mimic} to achieve 85\% training accuracy.
We utilize the efficient architecture known as ECAPA-TDNN~\cite{desplanques2020ecapa} to train this large classifier.
The penultimate layer embeddings (128-D) for a given PPG can be used as a representation of patient identity, which we leverage in our framework for domain adversarial learning~\cite{ganin2016domain} using a Gradient Reversal Layer (GRL).
We choose the supervised formulation from Additive Angular Margin (AAM)~\cite{deng2019arcface} to train our network, which we term the p-vector model, i.e., the patient identity vector model, and refer to its embeddings as simply p-vectors.
Although p-vector can also be learned through an unsupervised method via contrastive learning (positive = same patient, negative = otherwise), as in Apple PPG/ECG-FM~\cite{large-scale-training}, given the availability of identity labels in our dataset, we chose a supervised formulation to obtain more discriminative and higher-performing representations.
Note that PPG-based biometric identification has been addressed in prior works~\cite{luque2018end}; however, a large-scale identification model has not yet been demonstrated.

\noindent\textbf{Setup.}
For a minibatch of size $B$, let inputs $x_i$, CA labels $y_i\in\{0,1\}$, and pvectors $p_i\in\mathbb{R}^{d_p}$.
BLSTM $F(\cdot;\theta)$ yields $h_i=F(x_i;\theta)\in\mathbb{R}^{d_h}$
; a CA head $C(\cdot;\phi)$ outputs the logit $s_i=C(h_i;\phi)$ with $\hat y_i=\sigma(s_i)$; an adversary $A(\cdot;\psi)$ tries to
predict pvectors from a gradient-reversed feature: $\hat p_i=A(R_{\alpha}(h_i);\psi)$. The adversary is simply a linear layer appended to BLSTM.

\noindent\textbf{Gradient Reversal Layer.}

\begin{equation}
    R_\alpha : \mathbb{R}^{d_h}\to \mathbb{R}^{d_h}, 
\begin{cases}
R_\alpha(h) = h & \text{(forward)}, \\[6pt]
\nabla_h R_\alpha(h) = -\alpha I & \text{(backward)}.
\end{cases}
\end{equation}


\noindent\textbf{Adversarial pvector loss (MSE).}
\begin{equation}
\mathcal{L}_{\mathrm{pvec}}
=\frac{1}{B}\sum_{i=1}^{B}\big\|\,\hat p_i - p_i\,\big\|_2^2
=\frac{1}{B}\sum_{i=1}^{B}\big\|\,A\!\big(R_{\alpha}(h_i);\psi\big) - p_i\,\big\|_2^2 .
\end{equation}

\noindent\textbf{Total loss.}
\begin{equation}
\mathcal{L}_{\mathrm{total}}
=\mathcal{L}_{\mathrm{cls}}+\lambda_{\mathrm{adv}}\,\mathcal{L}_{\mathrm{pvec}} .
\end{equation}
$\mathcal{\lambda}_{\text{adv}}$ is set to 0.1 and $\alpha$ is set to 0.5 to control the strength of the adversary.

It is worth noting that patient identity information, while treated here as a confounder to suppress, can also be exploited in an alternative paradigm for \emph{model personalization}. We examined providing the patient identity vector explicitly to do zero-shot personalization~\cite{kataria2023self}, but it did not bring improvements. This can be explained as follows: First, our p-vectors, trained via ECAPA-TDNN with ID labels, capture identity signatures without incorporating information on cardiac health or arrest status. These embeddings capture identity signatures that may arise from multiple sources unrelated to cardiac status (e.g., sensor placement, demographic traits, or baseline waveform idiosyncrasies). Two patients with similar cardiac risk profiles may therefore have very different p-vectors. Providing such identity embeddings as explicit personalization factors, rather than treating them adversarially, would not benefit our model in capturing cardiac arrest–related features. Second, directly using ID embeddings risks severe overfitting during training by memorizing which IDs correspond to positive or negative outcomes, harming generalization at test time. Unlike true personalization paradigms, where the same patient’s signals are used in both training and testing, our framework does not support such a setup.
It is worth noting that patient leads and beds may change frequently in the ICU, leading to further challenges in personalization efforts.
Thus, adversarial suppression of identity is a more suitable approach than zero-shot personalization in this setting. Previous studies have reported similar issues, often addressing them by truncating identity embeddings into lower-dimensional spaces to reduce overfitting~\cite{senior2014ivector}.



\subsection{Pseudo-lab trajectory alignment}
Here, we propose leveraging auxiliary helper networks to produce estimates of blood biomarkers related to cardiac arrest, including lactate, sodium, troponin, and potassium~\cite{lott2021european,jones2023antecedents}.
Typically, EHR information includes time-sparse lab test results.
For datasets with sparse or no labels, we can use a proxy, referred to as pseudo-label values.
These auxiliary networks are all trained using approximately 143 hours of labeled data from the UCSF dataset.
The networks are obtained by fine-tuning PPG-GPT FM, and the corresponding performance for the true lab prediction results is in Table~\ref{tab:pseudo}.
We also refer to these networks as teacher models, which produce pseudo-labels on the fly during CA training.
The process can also be viewed as a proxy for \emph{aligning with the patient's true lab trajectory} (with interpolation to make it continuous).
Ideally, we need true values at high frequency, but we explore the middle ground where we use \emph{approximate values but at a high frequency}, i.e., every 30 seconds, thus using the term trajectory.
Since the prediction is far from perfect, the pseudo-lab values are noisy.
To facilitate learning from noisy objectives, we utilize the PCGrad~\cite{yu2020gradient} technique in our MTL framework.
The complete formulation is given below:

\begin{table}[htbp]
\centering
\begin{tabular}{l|c}
\hline
Blood biomarker  & MAE ($\downarrow$) \\ \hline
Lactate (mmol/L) & 1.1  \\ \hline
Sodium (mEq/L) & 6.85  \\ \hline
Troponin (ng/ml) & 6.71  \\ \hline
Potassium (mEq/L) & 0.5  \\ \hline
\end{tabular}
\caption{Blood biomarker prediction performance by auxiliary networks that were obtained by fine-tuning PPG-GPT on true labels. Thereafter, these are used to generate pseudo-values for CA PPGs for biomarker alignment.}
\label{tab:pseudo}
\end{table}

\noindent\textbf{On-the-fly pseudo labels for labs with PCGrad.}
For a batch $\{(x_i,y_i)\}_{i=1}^B$, let $y_i\in\{0,1\}$ be cardiac-arrest (CA).
We index laboratory auxiliaries by $\mathcal{L}=\{\text{lac},\text{na},\text{trop},\text{k}\}$ (lactate, sodium, troponin, potassium).
BLSTM $F(\cdot;\theta)$ yields $h_i=F(x_i;\theta)\in\mathbb{R}^{d_h}$.
A CA head $C(\cdot;\phi)$ outputs logit $s_i=C(h_i;\phi)$ with $\hat y_i=\sigma(s_i)$.
For each lab $\ell\in\mathcal{L}$ we use a \emph{distinct frozen teacher model} $T_\ell$ (forward only) to generate an on-the-fly pseudo label
\[
\tilde v_i^{(\ell)} \;=\;T_\ell(x_i)\quad\text{with }T_{\text{lac}},T_{\text{na}},T_{\text{trop}},T_{\text{k}}\text{ all different.}
\]
A student head $G_\ell(\cdot;\varphi_\ell)$ predicts $\hat v_i^{(\ell)}=G_\ell(h_i;\varphi_\ell)$.


\noindent\textbf{Auxiliary losses (per lab).}
\begin{equation}
\mathcal{L}_{\mathrm{lab}}^{(\ell)}
=\frac{1}{B}\sum_{i=1}^{B}\big\|\hat v_i^{(\ell)}-\tilde v_i^{(\ell)}\big\|_2^2,
\qquad \ell\in\mathcal{L}.
\end{equation}

\noindent\textbf{Total loss.}
\begin{equation}
\mathcal{L}_{\mathrm{total}}^{(\ell)}
=\mathcal{L}_{\mathrm{cls}}+\lambda_{\mathrm{lab}}\,\mathcal{L}_{\mathrm{lab}}^{(\ell)} .
\end{equation}


\noindent\textbf{PCGrad for gradient conflict removal.}
Compute per-task gradients $g_j=\nabla_{\Theta}\widetilde{\mathcal{L}}_j$
w.r.t.\ shared parameters $\Theta$.
For $i\neq j$, if $g_i^\top g_j<0$ (conflict) project
\begin{equation}
g_i \leftarrow g_i - \frac{g_i^\top g_j}{\|g_j\|_2^2+\varepsilon}\,g_j,
\end{equation}
else leave $g_i$ unchanged. The final update direction is the mean of the projected gradients:
\begin{equation}
g^\star=\frac{1}{|\mathcal{T}|}\sum_{i\in\mathcal{T}} g_i,\qquad
\Theta \leftarrow \Theta - \eta\,g^\star .
\end{equation}


Note that teacher models are frozen and $\lambda_{\text{lab}}$ is set to 2.

\section{Results}
\subsection{Baseline}
We first search for the best foundation model for our experiments~\cite{ni2025ppgdistillefficientphotoplethysmographysignals}, while the baseline model FEAN fixed it to PPG-GPT.
We report results with two metrics: AUROC (Area Under the Receiver Operating Curve) and AUPRC (Area Under the Precision-Recall Curve).
We typically report a single number (time-averaged AUC), which is the average of these metrics obtained at 24 lead times, i.e., T-25, ..., up to T-1, where T is the event time.
In Table~\ref{tab:baseline}, we compare 19M PPG-GPT with small variants of PaPaGei (5M) and MOMENT (40M).
These competing models are shown to be performant in many clinical and non-clinical tasks; however, PPG-GPT takes a strong lead due to its training on millions of hours of ICU data from the same domain, which also avoids the out-of-domain generalization issue.
Our study is the first to indicate the superiority of PPG-domain specialist FMs for challenging long-horizon prediction tasks.

\begin{table}[!t]
\centering
\begin{tabular}{l|c|c}
\hline
Foundation Model  & AUROC & AUPRC \\ \hline
PaPaGei & 0.566 & 0.148 \\ \hline
MOMENT & 0.660 & 0.252 \\ \hline
PPG-GPT & 0.736 & 0.315 \\ \hline
\end{tabular}
\caption{Baseline results by different Time-Series Foundation Models}
\label{tab:baseline}
\end{table}

\begin{figure*}[!t]          
  \centering

  \begin{subfigure}[b]{0.99\linewidth}
    \centering
    \includegraphics[trim=0 0 0 19,clip,width=\linewidth]{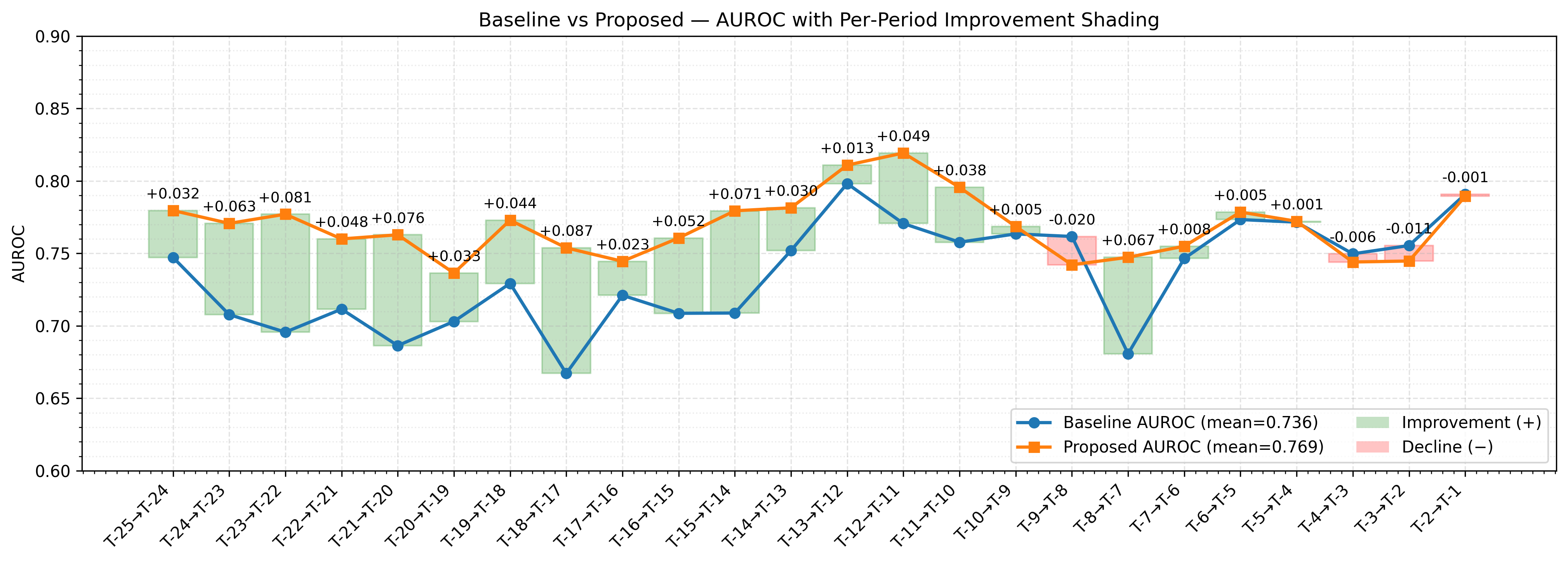}
    \caption{AUROC comparison}           
    \label{fig:tte_auroc}
  \end{subfigure}

  \vspace{1em}              

  \begin{subfigure}[b]{0.99\linewidth}
    \centering
    \includegraphics[trim=0 0 0 19,clip,width=\linewidth]{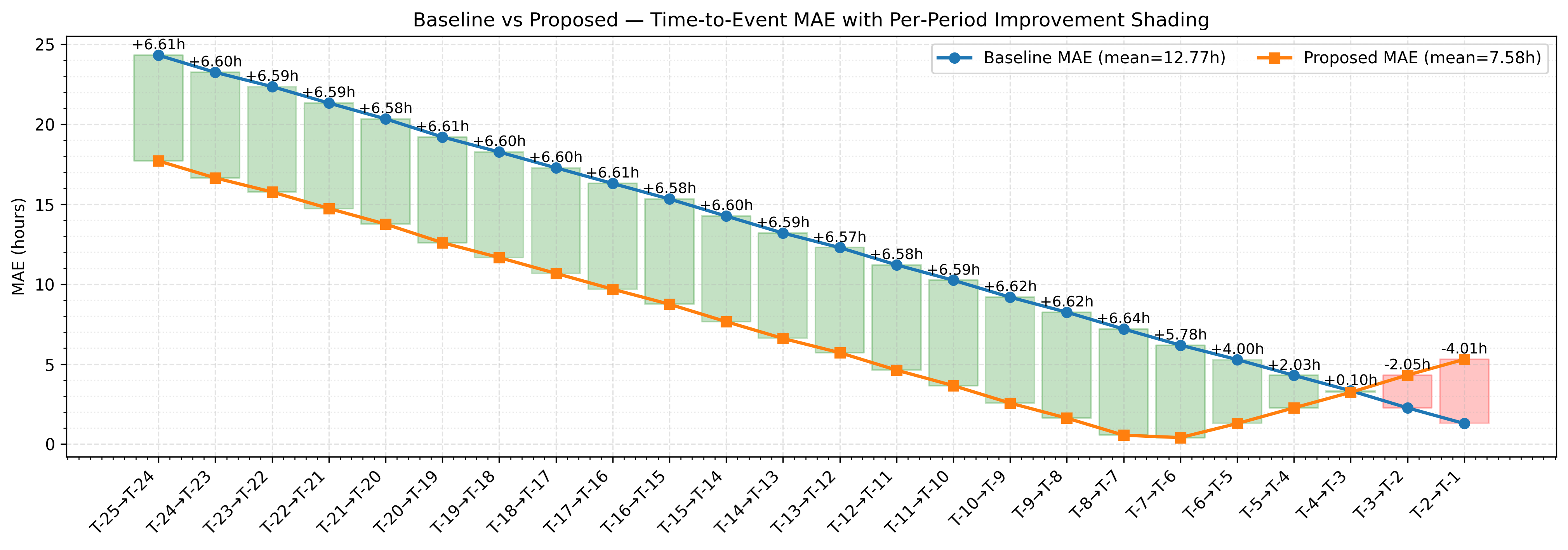}
    \caption{TTE-MAE comparison}
    \label{fig:tte_mae}
  \end{subfigure}

  \caption{Performance of baseline vs. proposed model.%
           (a) AUROC and (b) TTE-MAE.}  
  \label{fig:tte}
\end{figure*}

\subsection{Results with Time-to-event modeling}
The 24-hour time-averaged AUROC and AUPRC with TTE loss are in the second and third row of Table~\ref{tab:all}.
TTE modeling with simple regression loss (TTE-reg) yields an ROC of 0.783 and an AUPRC of 0.346.
The survival formulation (TTE-survival) yields an ROC of 0.730 and an AUPRC of 0.320, slightly outperforming the baseline in AUPRC.
The second formulation is significantly more complex, with a challenging experimental setup (small data, severe data imbalance, noisy labels).
Even hyperparameter tuning yields minimal benefits.
We want to emphasize that TTE-survival achieved a very low training error, which indicates potential over-fitting and a high potential for resource-rich tasks -- which we will investigate in future work.
Note that we also tried a simpler constant-effect hazard model, which yielded better performance than TTE-survival but not as good as TTE-reg.
This indicates that our setup prefers simpler formulations.

In Figure~\ref{fig:tte}, we show the improvement of TTE-reg for each lead time.
In the Figure~\ref{fig:tte_auroc}, we show AUROC with green shading indicating improvement and red indicating degradation.
We note that benefits primarily come from longer lead times, i.e., we improve prediction when the event is 12-24 hours away.
In Figure~\ref{fig:tte_mae}, we note the improvement in estimating TTE.
Our model exhibits consistent improvement from T-25 to T-8, after which the improvement becomes less pronounced and even worse than random prediction near the event. However, the AUROC remains unchanged, indicating robustness to wrong TTE estimates.

\begin{figure*}[b]
\centering
\includegraphics[trim=0 0 0 19,clip,width=0.95\linewidth]{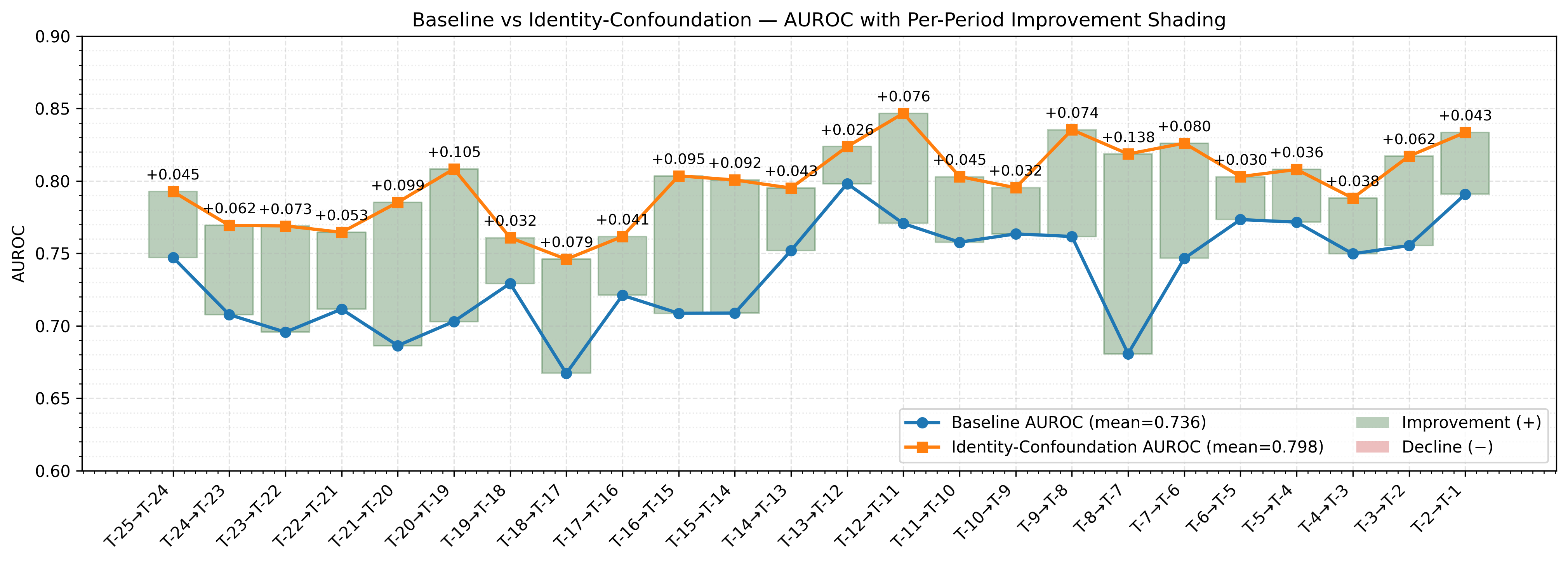}
\caption{Comparison of baseline with identity-invariant cardiac arrest model}
\label{fig:identity}
\end{figure*}

\begin{figure*}[!b]
\centering
\includegraphics[width=0.99\linewidth]{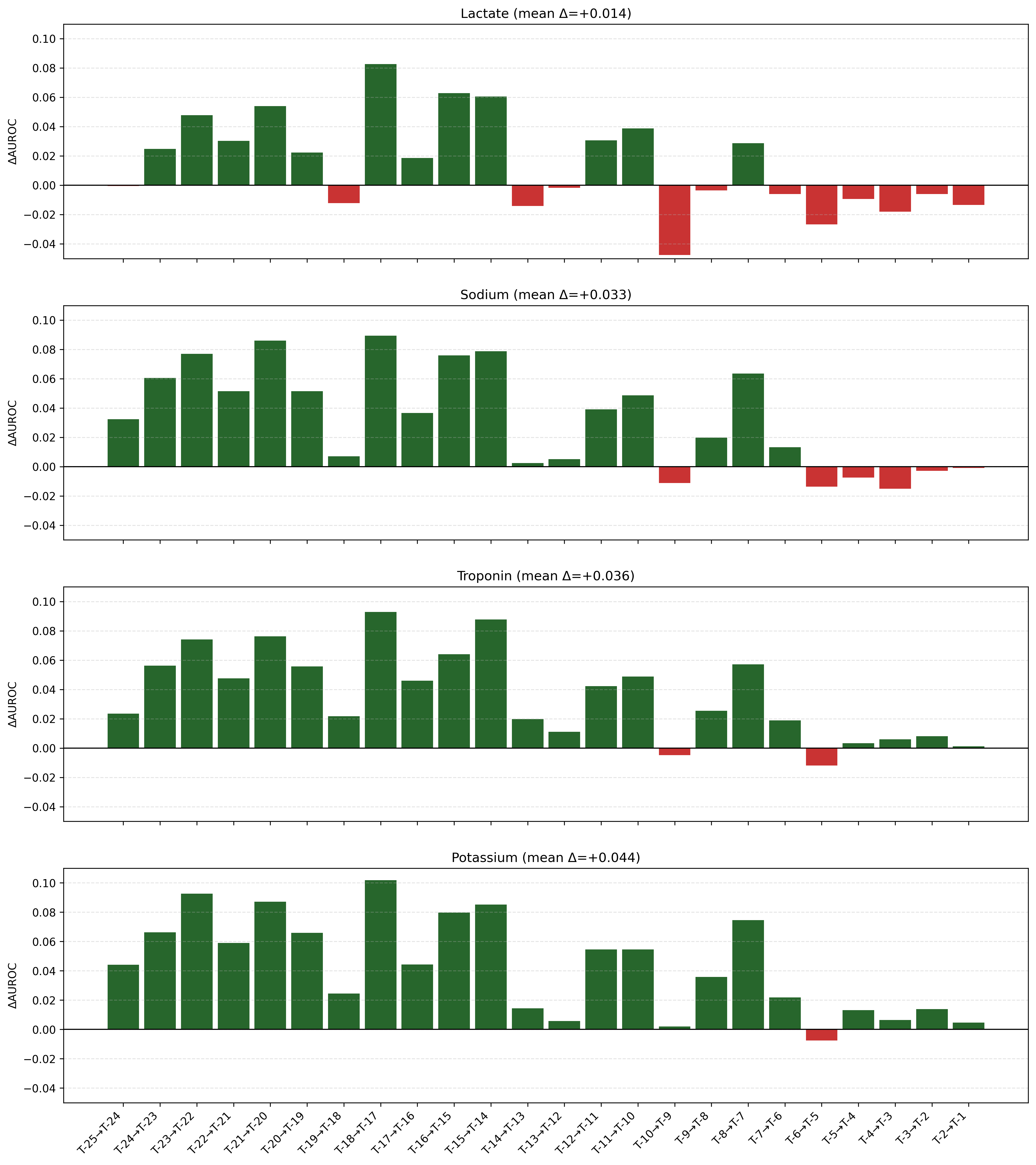}
\caption{Improvement in AUROC over baseline using different pseudo-lab teacher models}
\label{fig:pseudo}
\end{figure*}

\begin{table}[!b]
\centering
\begin{tabular}{c|l|c|c}
\hline
Index & Additional losses to classification  & AUROC & AUPRC \\ \hline
1 & None & 0.736 & 0.315 \\ \hline
2 & TTE-reg & 0.783 & 0.346 \\ \hline
3 & TTE-survival & 0.730 & 0.320 \\ \hline
4 & pvector & \textbf{0.798} & \textbf{0.350} \\ \hline
5 & pseudo(lac) & 0.750 & 0.278 \\ \hline
6 & pseudo(na) & 0.769 & 0.299 \\ \hline
7 & pseudo(trop) & 0.772 & 0.308 \\ \hline
8 & pseudo(k) & 0.779 & 0.303 \\ \hline
9 & TTE+pvector+pseudo(trop) & 0.770 & 0.340 \\ \hline
\end{tabular}
\caption{All results}
\label{tab:all}
\end{table}

\subsection{Results with identity invariance}
This idea yields the maximum benefit, as shown by the ``pvector'' row in Table~\ref{tab:all}.
The improvement is also shown in detail in Figure~\ref{fig:identity}.
In contrast to the TTE results, we observe benefits in all lead times, with slightly higher improvements in longer horizon windows (in absolute terms).
Note that for these results, we used an 85\% accuracy patient identification model with Out-of-domain (OOD) 7k subjects from MIMIC data.
Our preliminary studies show that the pvector model already captures other demographic information, such as gender and age.
We believe that stronger biometric models will be able to better avoid overfitting to patient identity and potentially other variables, such as ICU type, skin tone, and hospital hardware, paving the way to build \emph{fairer} models~\cite{panchumarthi2025fairtunebiasawarefinetuningframework}.

\subsection{Results with pseudo-lab alignment}
In Table~\ref{tab:all}, rows ``pseudo(lac)'', ``pseudo(na)'', ``pseudo(trop)'', and ``pseudo(k)'' show the average results with the potassium model giving the best results.
The improvement or degradation by lead time is shown in bar charts in Figure~\ref{fig:pseudo}.
Like TTE, we observe that most improvements are in regions with longer lead times, from T-25 to T-6.
We even observe some degradations closer to the event.
Troponin and potassium show the most consistent improvements, while sodium and, especially, lactate exhibit degradations close to the event.
Note that these results are conditional upon the performance of pseudo-lab teacher networks and do not reveal the importance of electrolyte for CA, as they were obtained using a small experimental setup and noisy pseudo labels.

\subsection{Results with all objectives}
A limitation in our study is the ability to use objectives together in a multi-task fashion.
Even with uncertainty weighting~\cite{kendall2018multi} and learnable weights~\cite{sener2018multi}, the performance can reach a maximum of 0.770 AUROC as noted in the last row of Table~\ref{tab:all}.
Prior work, such as Physiological Deep Learner~\cite{cherifa2021physiological}, has attempted MTL for vital forecasting, but our tasks have much more complexity.
We attribute our challenge to \emph{gradient conflict} due to our choice of highly heterogeneous tasks, which suffer from differing scales, noisy labels, and domain mismatch.
An extensive grid search for hyperparameters and weight schedule can alleviate our issue, but it is computationally prohibitive.
The following equations are used to compute conflict.

\begin{figure}[!t]          
  \centering
  \begin{subfigure}[b]{0.90\linewidth}
    \centering
    \includegraphics[width=\linewidth]{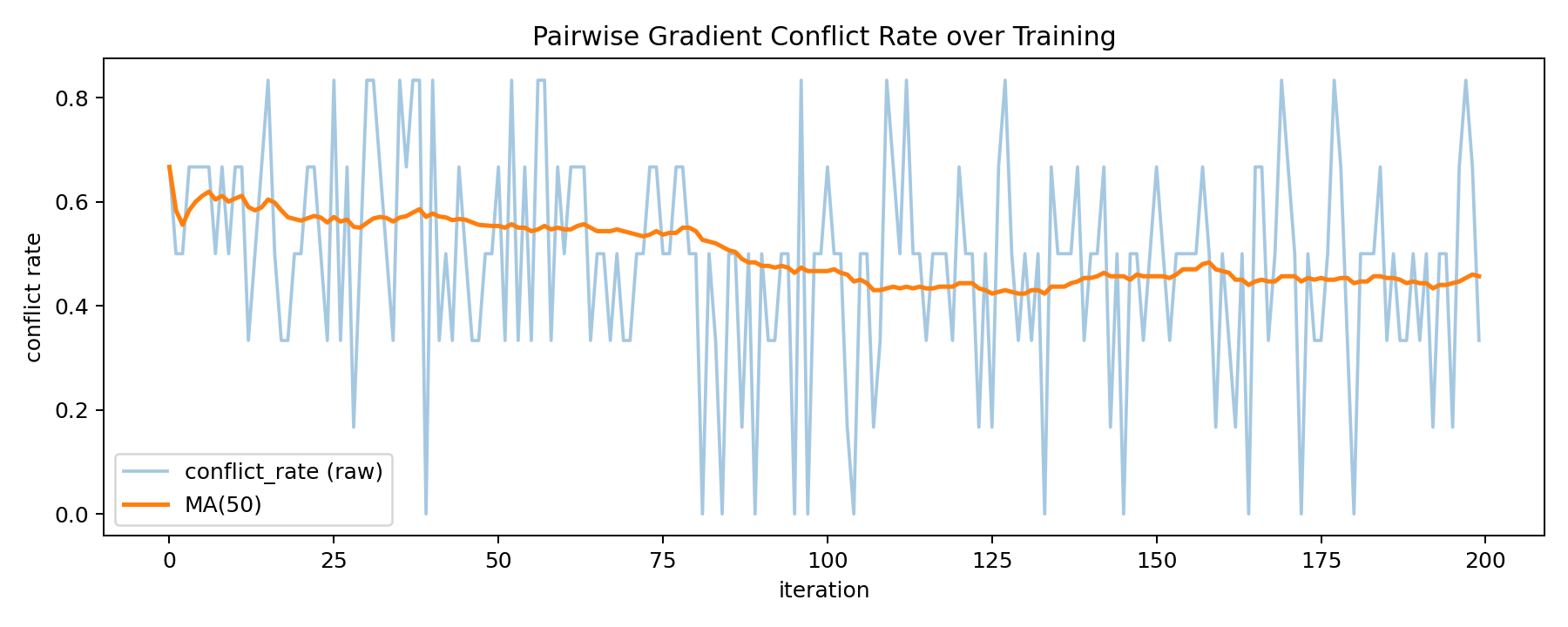}
    \caption{Overall conflict rate}           
    \label{fig:conflict_all}
  \end{subfigure}
  \vspace{1em}              
  \begin{subfigure}[b]{0.90\linewidth}
    \centering
    \includegraphics[width=\linewidth]{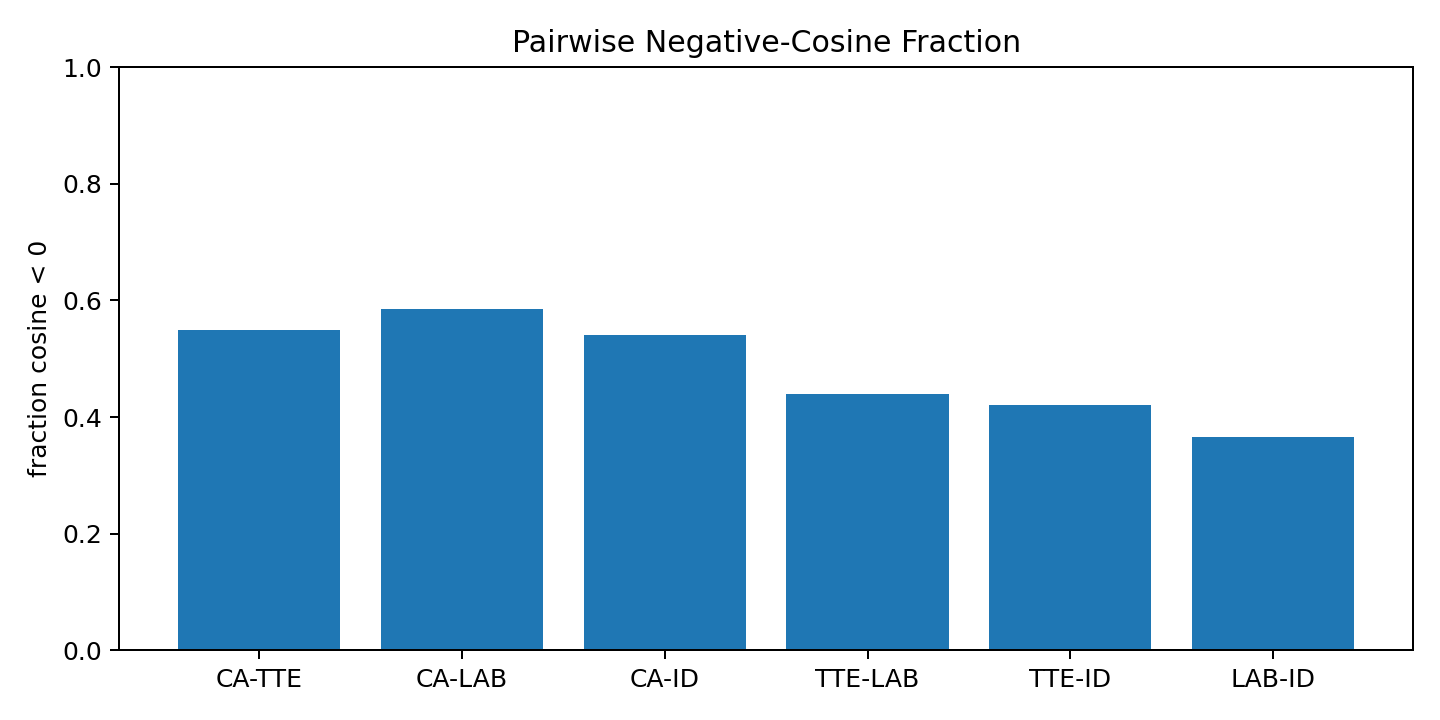}
    \caption{Intra-loss conflict rates}
    \label{fig:conflict_intra}
  \end{subfigure}
  \caption{Gradient conflict rate analysis for Multi-Task Learning optimization}  
  \label{fig:conflict}
\end{figure}

\noindent
\textbf{Tasks and gradients.}
Let $\mathcal{L}_{\text{CA}},\mathcal{L}_{\text{TTE}},\mathcal{L}_{\text{LAB}},\mathcal{L}_{\text{ID}}$ be the four losses and let
$\theta$ denotes the shared parameters. For task $i\in\{\text{CA},\text{TTE},\text{LAB},\text{ID}\}$ the stepwise gradient is
$\mathbf{g}_i=\nabla_\theta \mathcal{L}_i$.

\noindent
\textbf{Cosine similarity and conflict.}
The pairwise cosine is
\begin{equation}
s_{i,j}=\cos(\mathbf{g}_i,\mathbf{g}_j)=\frac{\mathbf{g}_i^\top \mathbf{g}_j}{\lVert \mathbf{g}_i\rVert_2\,\lVert \mathbf{g}_j\rVert_2}.
\end{equation}
A conflict occurs when $s_{i,j}<0$. With $K=4$ tasks, the conflict rate at step $t$ is
\begin{equation}
r_t=\frac{2}{K(K-1)}\sum_{i<j}\mathbb{I}\!\left[s_{i,j}^{(t)}<0\right].
\end{equation}

In Figure~\ref{fig:conflict_all}, we plot the conflict rate during training.
Note that it starts at approximately 60\% and then stabilizes to a minimum of 45\%.
The orange curve represents the moving average for a clearer interpretation.
In Figure~\ref{fig:conflict_intra}, we show the intra-conflict rate between the tasks.
All tasks have high rates among them, with the minimum (40\%) observed between pseudo-lab objective and identity invariance.
In the future, we plan to explore data augmentation as a means to mitigate some of the challenges associated with our current setup.

\section{Conclusion}
In this study, we proposed three ideas to improve low-resource PPG waveform-only cardiac arrest prediction.
Foundation models offer strong features, but their downstream performance depends on how they are harnessed with fine-tuning techniques and loss formulations.
Due to the low-resource challenge (i.e., a limited number of positive samples) in our setup, we devise auxiliary labels.
Through time-to-event modeling, we observed that a simple regression loss is a dependable choice, often outperforming hazard formulations and yielding improvements over long horizons.
Through enforcing patient identity invariance in cardiac arrest features via domain adversarial learning, facilitated by a large patient biometric model, we achieve the highest improvements.
By predicting pseudo-lab values generated by auxiliary networks, we align cardiac arrest features with the evolving underlying physiology of the patient and observe uniform benefits.
Finally, we analyze the multi-task learning formulation by examining the gradient conflict rate.
Our work contributes several ideas to improve the tracking of patient deterioration, particularly over long horizons.
In the future, we can mitigate low-resource constraints by harmonizing different data sources and conducting cross-institutional studies, or jointly pursuing other critical events, such as sepsis.

\bibliographystyle{elsarticle-num}
\bibliography{elsarticle.bib}
\end{document}